# Knowledge Retrieval

UJWAL SAINI
DEPARTMENT OF COMPUTER SCIENCE AND ENGINEERING
UNIVERSITY OF SOUTH FLORIDA
TAMPA, USA
SAINIU@USF.EDU

*Abstract*— We have designed three search methods for producing the task trees for the provided goal nodes using the Functional Object-Oriented Network. This paper details the strategy, the procedure, and the outcomes

## I. INTRODUCTION

The development of intelligent agents that can recognize human intentions and act to address problems in human-centered domains has attracted a lot of attention in the field of robotics research. Among these industries are those that deploy robots to assist the disabled and old, transport meals, and perform culinary duties. The primary difficulty in developing robots for human-centered areas, however, is in the variety of professions and the dynamic nature of the surroundings in which these machines would function. When it comes to robotic cooking, there are many different states to take into account when following a recipe, as well as a wide range of shapes, sizes, and ingredients. These components must be made by the formatter using the relevant criteria listed below. Additionally, there may be times when a robot cannot complete a dish because it is lacking certain items or ingredients in its surroundings; this may happen when a robot is required to prepare different meals.

The knowledge representation we use in this study is the Functional Object-Oriented Network (FOON), which builds on past work on combined object-action representation. In past work, we demonstrated how creating a FOON from video annotations might facilitate job planning. However, task planning is limited by the knowledge included in a FOON since, like to earlier efforts, it only gives knowledge for a limited number of recipe and ingredient modifications. For instance, there would be a problem if a robot created a salad with a particular combination of ingredients that had never been used together in FOON before since there is no known recipe for that particular type of salad.

We previously considered how FOON's knowledge may be expanded to encompass ideas from various object kinds. In light of this information, we propose that it is feasible to produce new, alternative solutions (as graphs) by utilizing the present understanding of analogous recipes. Since a reference task tree is received from FOON, we have built search methods in this work to generate task trees for specified target nodes.

## II. BACKGROUND

### A. Functional Object-Oriented Network

Motion nodes and object nodes are the two main types of nodes that make up a FOON, which is a bipartite network. Affordances are represented by edges that establish connections between things and actions and impose action sequencing. By describing the state change of objects before and after execution, a fundamental element called a functional unit—which contains input object nodes, output object nodes, and a motion node—depicts activities in FOON. Input and output nodes indicate preconditions and effects, respectively, similar to a planning operator in PDDL.

Videos of presentations are frequently annotated to create FOONs. An activity-specific FOON called a subgraph is made up of functional units that describe the circumstances of the objects before and after each action as well as the objects that are being handled. Although graph annotation is now done manually, past studies explored the possibility of semi-automatic annotation. Two or more subparagraphs can be combined to form what we refer to as a universal FOON. A universal FOON can contain several versions of recipes once it has been coupled with a number of information sources.

### B. Basics of a FOON

As was already said, a FOON has two different types of nodes. Technically speaking, this kind of graph is referred to as a bipartite network. In a FOON, object nodes that are modified in the environment or used to manipulate other object nodes are designated as NOs. Generally speaking, we only focus on things that are being actively used or acted upon in a particular activity. Another type of node that explains how these things are moved about is called a motion node (denoted as NM). Picking and putting, pouring, cutting, and stirring are a few examples of these actions. Bowls, cups, and box objects are examples of such containers. Since things can potentially include other objects, we can distinguish them depending on the ingredients they contain.

Between object nodes and motion nodes as well as between object nodes and motion nodes are connected in a FOON. Only when we transform our bipartite network into a one-mode projected graph for network analysis, as we did before in, can items be connected to objects. Our graph's edges are drawn from one node to the next in a pattern that results in the occurrence of a certain object-state outcome.

Identify applicable funding agency here. If none, delete this text box.

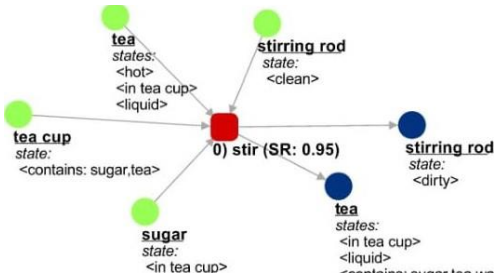

*C. The Functional Unit*

Individual, crucial learning units—what we refer to as functional units—make up a FOON. A subgraph is a group of functional units that constitute an activity with two or more phases, and a functional unit denotes a single, atomic action. In a film where the demonstrator is making macaroni and cheese, for instance, a subparagraph is created to show the entire activity. This paragraph might be made up of several units that explain procedures like adding macaroni pasta to boiling water, stirring it with a spoon, and then transferring the cooked pasta to a skillet.

## III. CREATING A FOON

Informational sources of knowledge, such as examples of human behavior or observations from instructional videos, are used to create a FOON. However, due to the challenges in differentiating the items being utilized, the states they are in, and the motion that is taking place, automated information extraction from such sources is incredibly tough. In the absence of such a system, we now annotate films manually; volunteers were tasked with choosing and annotating cooking videos. We also record the timestamps at which events occur in the source videos for future reference.

*A. Gathering and Combining Knowledge*

A selection of YouTube video sources served as the foundation for the knowledge represented by a FOON. A subgraph will be created for each source video in which the functional units are created by manual annotation. The annotation approach simply involves noting events that happen in movies, including their timing, the objects involved, any state changes, and the type of motion that takes place. The knowledge from these several subgraphs may then be combined into a single, larger FOON via a merging approach. The merging procedure is, in theory, very straightforward: we perform a union operation on all functional units while removing any duplicates. Duplication in this context denotes that input, object, and motion nodes in two units are completely identical.

*B. Universal FOON*

A combined collection of two or more subgraphs from various information sources is referred to as a universal FOON. A robot can utilize a universal FOON as a knowledge base to employ object-motion affordances to solve issues since it is composed of information from many sources. Our global FOON now consists of 65 YouTube source films that span a wide range of dishes. We provide download links on our website for all video subgraph files and samples of the FOON graphs discussed in this study for interested readers.

*C. . Knowledge Retrieval*

A robot will get information from a universal FOON that it may use to manipulate objects given a specified objective. A human user may provide instructions to a robot to make a meal within certain parameters. Finding a task tree—a collection of functional unit-based procedures that, when finished, fulfill a goal—is the aim of knowledge retrieval. A task tree is nothing more than a group of functional units that are most likely connected together and that, when carried out sequentially, act out the execution of actions that achieve a manipulation objective. Any FOON object node, whether it is a finished good or a thing in a transitional stage, can be this objective.

The retrieval method for a task tree sequence is based on the ideas of basic graph searching algorithms; when searching, we look into the depth of each functional unit, but specifically what ingredients or tools are in its immediate vicinity so that the system can determine whether or not a solution exists in that situation. The results of this search are either a task tree sequence (in which a target node is determined to be solvable and we have a functional unit sequence that creates the objective), no tree due to time limits, or no tree at all.

We employ the number of units (or steps) in our search as a heuristic for identifying the ideal task tree. The search method only considers the first unit that can be executed fully (or, more specifically, where all objects required are available as input to that unit). An object may consist of many units (for instance, different trees with/without the same step size), but the search method only considers the first unit that can be executed fully. Instead of using a step-based algorithm to find a tree, we may resolve ties in functional units depending on the difficulty of the jobs. Due to limitations in its configuration space or design, a robot could occasionally be unable to carry out a certain motion. However, we can make up for this by making a more straightforward change that produces the same outcomes.

## IV. METHODOLOGY

*A. Iterative Deepening Search*

Iterative Deepening Search (IDS) is an iterative graph searching approach that consumes substantially less memory in each iteration while benefiting from the completeness of the Breadth-First Search (BFS) strategy (similar to Depth-First Search). IDS accomplishes the needed completeness by imposing a depth limit on DFS, which reduces the danger of becoming stuck in an infinite or very long branch. It traverses each node's branch from left to right until it reaches the appropriate depth. After that, IDS returns to the root node and explores a separate branch that is comparable to DFS.

1)Time & space complexity: Assume we have a tree in which each node has b children. This will be our branching factor, and d will be the tree's depth. Nodes on the lowest level, $d$, will be extended exactly once, whereas nodes on levels $d-1$ will be expanded twice. Our tree's root node will be extended $d+1$ times. If we combine all of these terms, we get:

(d)b+(d−1)b2+...+(3)bd−2+(2)bd−1+b

Summation of time complexity will be: O(bd )

The space complexity is: O(bd), In this case, we suppose b is constant and that all children are formed at each depth of the tree and saved in a stack during DFS.

2) Performance analysis: Continually looping over the same nodes may give the impression that IDS has a substantial overhead, although this is untrue. This is because the algorithm only sometimes traverses a tree's lowest levels. The cost is kept to a minimum since upper-level nodes do not make up the bulk of nodes in a tree.

3) Implementation: To discover the best answer, we must examine every avenue. We simply followed the first trail we came across to keep things easy. Until we found the answer, we kept deepening the search. If the leaf nodes are available in the kitchen, the task tree is regarded as a solution.

*B. Greedy Best-First Search:*

For huge search spaces, the informed search algorithm is more useful. Because an informed search algorithm employs heuristics, it is also known as a Heuristic search.

**Heuristics function**: Heuristic is a function in Informed Search that finds the most promising path. It takes the agent's current state as input and calculates how close the agent is to the goal. The heuristic method, on the other hand, may not always provide the greatest solution, but it will always discover a good solution in a fair amount of time. The heuristic function calculates how close a state is to reach the goal. It is denoted by h(n), and it computes the cost of an optimal path between two states.

Heuristics 1: Here we are considering the motion rates for selecting the Input nodes as the heuristic function. The basic pseudocode follows-

```
Input: Given Goal node G and ingredients I
T ← A list of functional units in Task tree.

Q ← A queue for items to search
Kingd ← List of items available in kitchen.
Q.push(G)
While Q is not empty do:
    N ← Q.dequeue()
    If N not in Kingd then:
        C ← Find all functional units that create C
        max = -1
        for each candidate in C do:
            if candidate.successRate > max then:
                max = candidate.successRate
                CMax = candidate
        End for
        T.append(CMax)
        for each input in Cmax do:
            if n is not visted then:
                Q.enque(n)
                Make n visitied
            End if
        End for
    End if
End while
T.reverse()
Output: T
```

Heuristics 2: This algorithm is similar to the second in that the number of input nodes and their components are considered while selecting the candidate unit. A functional unit with the fewest input nodes will be picked as a candidate unit at each level.

### DISCUSSION

By running DFS and BFS at the selected depth bound, an iterative deepening search investigates the FOON. There won't be a solution, thus the depth level will keep increasing. This method takes more time to build the task tree if the solution appears at a deeper level. Revisiting all previously visited nodes for each depth-bound increment, will increase the temporal complexity. Heuristics 1 and 2 can readily discover the solution at higher levels since they follow BFS, however, each complexity rises if the solution appears at deeper layers.

| Goal Nodes | IDS | Heuristics 1 | Heuristic 2 |
|---|---|---|---|
| Greek Salad | 31 | 32 | 28 |

| Ice | 1 | 1 | 1 |
|---|---|---|---|
| Macaroni | 7 | 7 | 8 |
| Sweet potato | 3 | 3 | 3 |
| Whipped Cream | 10 | 10 | 15 |

The task trees for all three methods could have the same or different numbers of functional units. All task trees have the same number of functional units for the target nodes ice and sweet potato